\documentclass[11pt]{article}

\usepackage{acl}

\usepackage{times}
\usepackage{latexsym}
\usepackage[LGR, T1]{fontenc}
\usepackage[greek, english]{babel}
\usepackage{alphabeta}
\usepackage[utf8]{inputenc}
\usepackage{microtype}
\usepackage{inconsolata}

\usepackage{graphicx}
\usepackage{booktabs} 
\usepackage{multirow}
\usepackage{float}

\usepackage{subcaption} 
\usepackage{enumitem}

\title{Pass or Fail? Evaluating LLMs on Two Greek Examination Benchmarks}

\author{{Panagiota Kyriazi, Eleni Kasoura, Prokopis Prokopidis} \\
  Institute for Language and Speech Processing / Athena RC \\
  \texttt{\{p.kyriazi, eleni.kasoura, prokopis\}@athenarc.gr}}

\begin{document}

\maketitle

\begin{abstract}
The rapid advancement of Large Language Models (LLMs) imposes a thorough evaluation of their linguistic and analytical capabilities as well as constraints, particularly for a language with limited benchmark coverage such as Greek. To address the limited availability of comprehensive benchmarks in this domain, we introduce \textbf{Prot-Ex} and \textbf{Pan-Ex}, two benchmarks consisting of questions from entrance exams for Greek Model and Experimental schools as well as the Panhellenic exams (the Greek national university entrance examinations). These benchmarks are employed to assess the performance of text-only LLMs—including the Greek-adapted KriKri-8B-Instruct, Llama-3.1-8B, Gemma-4-26B, and Qwen-3-32B—across diverse academic disciplines (Modern Greek, Mathematics, Physics, etc.) and task formats (closed, structured, and open-ended), including textualized visual context (i.e., image descriptions). Our findings indicate the localized KriKri-8B significantly outperforms its base model, successfully rivalling much larger LLMs in linguistically demanding humanities tasks. By leveraging an LLM-as-a-Judge methodology, we expose the inadequacy of traditional lexical metrics for evaluating complex reasoning. Crucially, we uncover a few-shot prompting paradox: while synthetic examples improve accuracy in closed-ended questions, they severely overload the context window of 8B models in structured tasks, causing significant performance degradation. Ultimately, this study suggests targeted linguistic adaptation offsets lower parameter counts in specialized domains, despite the fragility of smaller models to prompt verbosity.
\end{abstract}

\section{Introduction}

Evaluation tasks are crucial tools in the field of natural language processing (NLP) to keep track of the progress in machine learning and communicate the potential issues, biases, or needs that may arise from these evaluations \citep{biderman2024lessonstrenchesreproducibleevaluation}. To enable this process, benchmarks are deployed as reference points for Large Language Models (LLMs) to assess their performance on an equal footing for comparison purposes. These mainly consist of one or multiple datasets along with relevant metrics and a standardized methodology to evaluate their performance \citep{ruder2021nlpbenchmarking}. 

Well-known benchmarks, such as General Language Understanding Evaluation (GLUE) \cite{wang-etal-2018-glue} and Cross-lingual TRansfer Evaluation of Multilingual Encoders (XTREME) \citep{hu2020xtrememassivelymultilingualmultitask}, are collections of tasks used for training, evaluation, and analysis of monolingual and/or multilingual language models in a series of NLU and NLP tasks. These benchmarks serve as an objective lens to observe each model's suitability for a specific task as well as the field's progression through remarkable findings.

However, issues regarding the transparency and reproducibility of LLM evaluations conducted by independent researchers have been noted. Specifically, concerns related to data contamination, where models may have seen test data during training, pose a major challenge to the validity of results \citep{sainz-etal-2023-nlp, zhou2023dontmakellmevaluation}. This highlights the need for a unified framework for result reproduction and novel evaluations on any LLM using already supported benchmarks \citep{biderman2024lessonstrenchesreproducibleevaluation, siddiq2025largelanguagemodelssoftware}. 

To this end, \citet{eval-harness} built the Language Model Evaluation Harness---known as \texttt{lm-eval}---which serves as an open-source research library for LLM evaluation. This infrastructure is publicly available, offering a unified framework to test generative language models on a large number of different evaluation tasks and compare the results across models in a standardized way.

In addition to \texttt{lm-eval}, the Inspect AI framework \citep{inspect_ai_2024} is a core evaluation platform, developed and open-sourced by the UK AI Security Institute (UK AISI). This tool aims to test the capabilities and security of frontier LLMs as well as to evaluate open-ended and complicated tasks, requiring human reasoning and critical thinking. It is publicly available, and it provides reproducible and structured evaluations by enabling the user to handle its infrastructure and create custom functions based on the task needs. 

Building upon the need for standardized assessment in non-English contexts, this project introduces the creation and evaluation of two novel Greek benchmarks: \texttt{greek-protipa-exams} (hereafter \textbf{Prot-Ex}) and \texttt{panhellenic-exams} (hereafter \textbf{Pan-Ex}). These datasets comprise questions and answers from official examinations for admission to model and experimental schools, as well as universities in Greece, respectively. Evaluations were conducted using both the \texttt{lm-eval} and Inspect AI frameworks. 

Protipa exam topics, as published online by the Greek Government from 2013 to 2026, include topics related to the subjects of Greek Language, Mathematics, Physics, and Religious Studies. Moreover, they cover levels of secondary education; specifically, the topics are divided into middle school and high school entrance exams.

The Panellinies dataset includes exam topics from 2020 to 2026, and the involved subjects are the following: Ancient Greek, Mathematics, Biology, Chemistry, Economics, Computer Science, History, Latin, Physics, and Greek Language. It consists of questions originating only from the general high school exams, offering a broad academic curriculum while preparing students for university entrance.

By leveraging the benefits of the lm-eval infrastructure combined with the Inspect AI framework, we evaluate the answers given by Llama-KriKri-8B-Instruct \citep{roussis-etal-2025-krikri} along with three other LLMs on both benchmarks to address the following research questions:
\begin{itemize}
\item \textbf{RQ1:} How does LLM performance differ (i) between open-ended and closed-ended tasks within the same subject and (ii) across different subjects?
\item \textbf{RQ2:} Are the classic evaluation metrics, such as BERTScore, reliable in comparison with more up-to-date LLM-as-a-Judge approaches when evaluating Greek educational data?
\item \textbf{RQ3:} In which type of tasks does the exploitation of few-shot examples have the most significant effect as opposed to baseline performance?
\end{itemize}

\section{Related Work}

LLM evaluation has evolved from early multimodal question-answering datasets to large-scale multi-subject suites such as MMLU \citep{hendrycks2021measuringmassivemultitasklanguage}, which benchmarks zero- and few-shot performance across 57 academic subjects spanning STEM, humanities, and social sciences, establishing a widely adopted standard for broad academic assessment.

While a plethora of multilingual question-answering benchmarks like Belebele \citep{Bandarkar_2024} and XNLI \citep{conneau-etal-2018-xnli} exist, their scope, question formats, and multimodal resources for the Greek language remain limited. Existing Greek benchmarks focus on tasks that include dialect identification \citep{chatzikyriakidis2025grdddatasetgreekdialectal}, Sign Language Translation \citep{voskou2023newdatasetendtoendsign}, and Ancient to Modern Greek machine translation \citep{mavromatis-etal-2026-ancient}, alongside speech processing evaluation resources spanning podcasts and regional dialects \citep{paraskevopoulos24_interspeech, tsoukala-etal-2026-extending}.

The ecosystem has been steadily expanding to encompass physical commonsense reasoning within broad cross-lingual initiatives such as Global PIQA \citep{mrl-workshop-2025-global-piqa}, domain-specific and educational evaluation suites covering medical exam datasets \citep{medical_mcqa_greek}, financial NLP with Plutus \citep{peng-etal-2025-plutus}, legal reasoning with GreekBarBench \citep{chlapanis-etal-2025-greekbarbench}, social-media-based QA with DemosQA \citep{mastrokostas2026evaluatingmonolingualmultilinguallarge}, empathetic support conversations for exam stress \citep{kyriazi-etal-2026-empathy}, and broad multi-subject multiple-choice evaluation through native-sourced GreekMMLU \citep{zhang-etal-2026-greekmmlu}. 

While GreekMMLU provides an excellent breadth of disciplines, it is confined to the multiple-choice format, measuring only discriminative accuracy. To contribute to the ecosystem and address this gap, we introduce the Prot-Ex and Pan-Ex benchmarks, derived from Greek Model and Experimental schools, as well as the national Panhellenic entrance exams. These datasets capture the rigorous nature of the national secondary education curriculum and extend beyond traditional closed-ended questions (CEQs) by featuring structured queries (SQs) and demanding open-ended generation tasks (OEQs) across diverse disciplines (e.g., Modern Greek, Mathematics, Physics). Furthermore, to support models in reasoning over visual information, we provide supplementary textualized visual contexts (image descriptions and transcriptions) for geometry diagrams and scientific figures. Finally, by integrating these benchmarks into established infrastructures like \texttt{lm-eval} and Inspect AI---and deploying an LLM-as-a-Judge pipeline to evaluate complex reasoning---we ensure transparent, standardized, and easily reproducible evaluation pipelines addressing the methodological limitations of prior studies.

\section{Methodology}

\subsection{Data Collection and Processing}
The data collection and processing tasks were identical for both datasets; we gathered the corpora of the publicly available exams of the experimental school and Panhellenic topics and organized them according to subject and year. To ensure consistency, raw files were renamed using standardized conventions, which were subsequently used for generating unique IDs for each QA pair. 

Each exam entry was processed and converted into a structured format. Specifically, questions were parsed into JSON files, while the corresponding solutions were extracted into Markdown files. The resulting datasets preserve essential information for LLM training and evaluation through the following keys: id (unique identifier), question (task description), input (passages or context), choices (candidate answers for closed tasks), answer (the ground truth), image (path and metadata for multimodal entries), and mark (grading score).

It is important to mention that the \texttt{image\_description} and \texttt{image\_transcription} fields were LLM-generated, and specifically, by Gemini 3.1 Pro. The images were provided as input to the LLM, along with an instructional prompt to analytically describe each image and transcribe any visual input that consisted of text. Subsequently, the generated descriptions and transcriptions were manually checked by the authors to ensure accuracy. This process was designed to assist non-multimodal LLMs in comprehending visual elements, providing them with the necessary textualized context in order to give the correct answer (see Appendix~\ref{sec:appendix_multimodal}).

\subsection{Dataset Statistics and Taxonomy}
To ensure rigorous evaluation and prevent data contamination, both benchmarks are partitioned into publicly accessible subsets and withheld private test sets.

The Prot-Ex dataset comprises a total of 1,766 entries, out of which 64 entries from the 2019 examinations are retained as a private evaluation set. We classified the total entries into two primary categories based on the required response type:
\begin{itemize}
\item CEQs (1,484 items): This category includes Multiple-Choice, True/False, and Fill-in-the-gaps.
\item OEQs (282 items): This category consists of Open questions, matching, and Fill-in-the-gaps tasks without provided choices.
\end{itemize}

The Pan-Ex dataset comprises a total of 1,540 entries, with 222 entries from the recent 2026 examinations strictly withheld to serve as a private evaluation set. The total entries are divided into the two respective categories as the aforementioned dataset:
\begin{itemize}
\item CEQs (454 items): This category includes Multiple-Choice and True/False tasks.
\item OEQs (1,086 items): This category consists of Open questions, matching, and Fill-in-the-gaps tasks without provided choices.
\end{itemize}

The distribution varies across subjects: most of the subjects include all task types, and Physics from protipa exams contains only open questions, while Religious Studies consists solely of closed (multiple-choice) items. A detailed subject-wise breakdown of these task formats for the public subsets is provided in Appendix~\ref{sec:appendix_public_distribution} and for the private in Appendix~\ref{sec:appendix_private_distribution}. Furthermore, the dataset supports multimodality, indicating entries that require visual context (diagrams, geometric shapes, maps) for their resolution, including text-based descriptions and transcriptions.

\begin{table*}[t]
    \centering
    \resizebox{\textwidth}{!}{%
    \begin{tabular}{l ccc ccc ccc ccc}
        \toprule
        \multirow{2}{*}{\textbf{Subject}} & \multicolumn{3}{c}{\textbf{KriKri-8B}} & \multicolumn{3}{c}{\textbf{Llama-8B}} & \multicolumn{3}{c}{\textbf{Gemma-4-26B}} & \multicolumn{3}{c}{\textbf{Qwen3-32B}} \\
        \cmidrule(lr){2-4} \cmidrule(lr){5-7} \cmidrule(lr){8-10} \cmidrule(lr){11-13}
        & \textbf{Closed} & \textbf{Structured} & \textbf{Open-ended} & \textbf{Closed} & \textbf{Structured} & \textbf{Open-ended} & \textbf{Closed} & \textbf{Structured} & \textbf{Open-ended} & \textbf{Closed} & \textbf{Structured} & \textbf{Open-ended} \\
        \midrule
        Greek Language    & 56.45 & 21.67 & 77.54 & 51.08 & 7.08 & 56.15 & 75.94 & 35.28 & 75.90 & 68.28 & 30.35 & 74.75 \\
        Mathematics       & 30.88 & -- & 66.89 & 30.72 & -- & 41.06 & 56.26 & -- & 87.02 & 52.09 & -- & 88.91 \\
        Physics           & -- & -- & 69.44 & -- & -- & 47.22 & -- & -- & 75.00 & -- & -- & 69.44 \\
        Religious Studies & 77.78 & -- & -- & 72.22 & -- & -- & 78.89 & -- & -- & 80.00 & -- & -- \\
        \midrule
        \textbf{Aggregate} & \textbf{47.10} & \textbf{21.67} & \textbf{69.93} & \textbf{43.89} & \textbf{7.08} & \textbf{45.48} & \textbf{67.90} & \textbf{35.28} & \textbf{83.46} & \textbf{62.25} & \textbf{30.35} & \textbf{84.21} \\
        \bottomrule
    \end{tabular}}  
    \vspace{-0.2cm}
    \caption{Zero-shot performance comparison (\%) for the Prot-Ex benchmark. We use the LM-Eval framework for the closed and structured question formats, and Inspect AI for open-ended questions. \textit{Note: The aggregate scores for Structured tasks reflect only the Greek Language subject, as other subjects do not contain questions in this format.}}
    \label{tab:rq1_protipa_results}
\end{table*}

\begin{table*}[t]
    \centering
    \resizebox{\textwidth}{!}{%
    \begin{tabular}{l ccc ccc ccc ccc}
        \toprule
        \multirow{2}{*}{\textbf{Subject}} & \multicolumn{3}{c}{\textbf{KriKri-8B}} & \multicolumn{3}{c}{\textbf{Llama-8B}} & \multicolumn{3}{c}{\textbf{Gemma-4-26B}} & \multicolumn{3}{c}{\textbf{Qwen3-32B}} \\
        \cmidrule(lr){2-4} \cmidrule(lr){5-7} \cmidrule(lr){8-10} \cmidrule(lr){11-13}
        & \textbf{Closed} & \textbf{Structured} & \textbf{Open-ended} & \textbf{Closed} & \textbf{Structured} & \textbf{Open-ended} & \textbf{Closed} & \textbf{Structured} & \textbf{Open-ended} & \textbf{Closed} & \textbf{Structured} & \textbf{Open-ended} \\
        \midrule
        Ancient Greek     & 51.11 & 21.04 & 57.06 & 51.11 & 20.00 & 35.31 & 68.89 & 46.04 & 65.53 & 75.56 & 28.75 & 64.62 \\
        Economics         & 54.76 & --   & 56.71 & 40.48 & -- & 24.32 & 78.57 & -- & 76.78 & 59.52 & -- & 74.59 \\
        Physics           & 48.65 & --   & 51.37 & 41.89 & -- & 21.23 & 66.22 & -- & 70.55 & 55.41 & -- & 74.66 \\
        History           & 63.33 & --   & 68.98 & 53.33 & -- & 44.44 & 50.00 & -- & 66.20 & 53.33 & -- & 63.33 \\
        Latin             & 73.33 & 12.31 & 51.68 & 66.67 & 6.15 & 32.72 & 66.67 & 15.38 & 64.03 & 73.33 & 15.38 & 66.14 \\
        Mathematics       & 54.84 & --   & 61.47 & 61.29 & -- & 19.47 & 87.10 & -- & 91.58 & 70.97 & -- & 93.42 \\
        Greek Language    & 96.67 & --   & 82.02 & 83.33 & -- & 56.55 & 86.67 & -- & 82.02 & 86.67 & -- & 79.88 \\
        Computer Science  & 56.67 & 7.50  & 71.35 & 56.67 & 2.50 & 41.35 & 93.33 & 12.50 & 77.88 & 86.67 & 12.50 & 86.54 \\
        Chemistry         & 48.08 & --   & 55.33 & 40.38 & -- & 23.17 & 76.92 & -- & 84.47 & 63.46 & -- & 88.50 \\
        Biology           & 51.35 & 20.48 & 63.38 & 32.43 & 16.90 & 30.30 & 56.76 & 35.00 & 69.75 & 56.76 & 54.52 & 70.86 \\
        \midrule
        \textbf{Aggregate} & \textbf{56.74} & \textbf{15.57} & \textbf{59.33} & \textbf{49.48} & \textbf{11.89} & \textbf{30.81} & \textbf{72.54} & \textbf{28.70} & \textbf{74.01} & \textbf{66.06} & \textbf{23.86} & \textbf{75.49} \\
        \bottomrule
    \end{tabular}}
    
    \vspace{-0.2cm}
    \caption{Zero-shot performance comparison (\%) for the Pan-Ex benchmark. We use the LM-Eval framework for the closed and structured question formats, and Inspect AI for open-ended questions.}
    \label{tab:rq1_panellinies_results}
\end{table*}

\subsection{Evaluation Setup}
For the evaluation process, we utilized the lm-eval-harness \citep{eval-harness}  and the inspect-ai \citep{inspect_ai_2024} frameworks. We used default settings and parameters for lm-eval-harness, along with a temperature of 0.1, a top-p of 0.9, and a top-k of 40 for inspect-ai in all evaluations described below. 

We assessed the performance of four LLMs:

\begin{enumerate}
\item Llama-KriKri-8B-Instruct \citep{roussis-etal-2025-krikri}: A model fine-tuned specifically for the Greek language.
\item Llama-3.1-8B \citep{grattafiori2024llama3herdmodels}: The foundation model of the Greek-focused Llama-KriKri-8B-Instruct, with the same size to compare their capabilities.
\item Gemma-4-26B \citep{gemmateam2026gemma4technicalreport}: An efficient instruction-tuned multimodal model with 25.2B total parameters from the Google DeepMind Gemma.
\item Qwen3-32B \citep{qwen3technicalreport}: A 32.8B parameter language model from the Qwen3 series, optimized for both complex reasoning and efficient dialogue.
\end{enumerate}

For accuracy purposes, we designed three different task configurations within the evaluation harness:

\begin{itemize}
\item CEQ Tasks: Evaluated using the lm-eval harness framework. The evaluation prompt explicitly instructs the model to output a specific identifier (e.g., A, B, 0, 1) corresponding to the correct choice. Performance is strictly measured using Exact Match accuracy.

\item SQ Tasks (Structured): This category consists of questions requiring constrained outputs—such as a single word or short phrase rather than extended generation. Evaluation is conducted via lm-eval utilizing a custom targeted metric (structured accuracy). To prevent false negatives caused by model verbosity, raw outputs undergo rigorous normalization (e.g., stripping Markdown syntax, filtering newlines, and truncating explanatory text) before being evaluated against the ground truth using format-specific matching rules.

\item OEQ Tasks: Deploys generative prompts that require the model to produce comprehensive, full-text responses or detailed reasoning. To overcome the limitations of traditional string-matching metrics, we employ a dual-evaluation strategy. First, BERTScore is utilized to capture semantic similarity against the reference answers. Second, the Inspect AI framework is deployed to implement an LLM-as-a-Judge evaluation paradigm, where we specifically used the Gemma-3-27b-it model as the grader.
\end{itemize}

\noindent \textbf{Computational Infrastructure and Compute Footprint:} All evaluation experiments were executed across a distributed computing setup: a local GPU server (NVIDIA GB10 GPU) for the 8B models and Gemma-3-27B-it judge scoring, a European HPC provider (NVIDIA A100 nodes) for closed/structured evaluation runs, a \$20 OpenRouter budget for Qwen3-32B and Gemma-4-26B, and Gemini 3.1 Pro for visual context preprocessing. Across all zero-shot and few-shot evaluation passes, total compute overhead is estimated at approximately 24 GPU hours.

\subsection{Data Availability Limitations}
During the data collection phase, certain inconsistencies were encountered with the source exam files. For the Prot-Ex benchmark, all files from 2015 were unavailable or corrupted, resulting in a gap in the dataset. Additionally, no solutions were provided for the Greek language subject for high schools in 2014, nor for the subjects of Greek and Math in 2018. Moreover, essay components of the Greek Language exams were excluded from the main benchmarks as no official answers are provided. 

Regarding the Pan-Ex benchmark, the 
solutions were published by the OEFE (Federation of Private Education Tutors of Greece) and are 
made publicly available as part of this dataset
for scientific and research purposes.

To mitigate potential data leakage for future community evaluations, the publicly released versions of our datasets exclude the 2026 data from Pan-Ex and the 2019 data from Prot-Ex. Consequently, our primary experimental results reported in Section~\ref{sec:results} are evaluated on these public datasets. To assess whether model performance remains consistent on unseen data, we additionally conducted experiments on the withheld ``private'' test sets (2019 for Prot-Ex and 2026 for Pan-Ex), with results reported in Appendix~\ref{sec:appendix_private}. Overall model rankings and relative task-format performance patterns on the private test sets closely mirror the public evaluation findings.

\section{Experimental Results and Analysis}
\label{sec:results}
We present the evaluation results for the involved LLMs on the Prot-Ex and Pan-Ex benchmarks. The analysis is structured around the three research questions defined in the introduction, aiming to assess the models' capabilities across different task formats, evaluation metrics, and in-context learning. 

\subsection{Comparative Performance Across Exercise Types and Subjects}
As depicted in Tables~\ref{tab:rq1_protipa_results} and~\ref{tab:rq1_panellinies_results}, the performance of the evaluated LLMs varies significantly based on both the task modality and the specific academic subject. The evaluation spans two distinct benchmarks: the Prot-Ex benchmark, focusing on four core subjects, and the expansive Pan-Ex benchmark, covering ten diverse disciplines across humanities and STEM. The tasks across these benchmarks are categorized into closed, structured, and open-ended generation.

Addressing the first sub-question (i) regarding performance differences between task types within the same subject, a clear pattern emerges across both benchmarks where SQ tasks consistently yield the lowest performance for all models. 

In the Prot-Ex benchmark, models unexpectedly score higher in OEQ tasks compared to CEQs in specific subjects. In Mathematics, Qwen3-32B achieves 88.91\% in OEQs versus 52.09\% in CEQ tasks. This may reflect the capacity of larger models to articulate reasoning steps correctly, even if they struggle to map their output to a strict multiple-choice format. Furthermore, the dataset design inherently dictates task availability; Religious Studies consists purely of CEQ items—where models demonstrate high knowledge recall (e.g., Qwen scoring 80.00\%)—while Physics contains solely OEQs.

Focusing on the Pan-Ex benchmark, the relationship between closed and open-ended performance is highly subject-dependent. For instance, in Ancient Greek, Qwen3-32B scores 75.56\% in CEQ tasks and 64.62\% in OEQs, but drops significantly to 28.75\% in SQ tasks. Conversely, in the Greek Language subject, performance in CEQ tasks is exceptionally high (KriKri scoring 96.67\%) and generally exceeds open-ended scores (82.02\%).

Concerning the second sub-question (ii) regarding performance across different subjects, larger parameter models (Gemma-4-26B and Qwen3-32B) predictably outperform the 8B models (KriKri and Llama) in aggregate, particularly in subjects demanding 
technical and scientific reasoning. 

In the Prot-Ex benchmark, the localized KriKri-8B demonstrates a competitive edge in the Greek Language subject. In OEQ tasks, KriKri-8B scores 77.54\%, outperforming both the larger Gemma-4-26B (75.90\%) and Qwen3-32B (74.75\%) models, reinforcing the importance of targeted 
training.

In the Pan-Ex benchmark, Qwen3-32B achieves outstanding open-ended scores in STEM subjects such as Mathematics (93.42\%), Chemistry (88.50\%), and Computer Science (86.54\%), establishing a substantial gap over its 8B counterparts. Furthermore, Gemma-4-26B records exceptionally high closed-ended scores in Computer Science (93.33\%) and Mathematics (87.10\%). However, a significant exception is observed in the Greek Language subject, where KriKri-8B achieves the highest closed-ended score (96.67\%) and ties with Gemma in OEQs (82.02\%). This highlights the profound impact of language-specific adaptation over sheer parameter count when processing linguistically demanding humanities subjects.

\begin{table*}[t]
    \centering
    \resizebox{\textwidth}{!}{%
    \begin{tabular}{l cc cc cc cc}
        \toprule
        \multirow{2}{*}{\textbf{Subject}} & \multicolumn{2}{c}{\textbf{KriKri-8B}} & \multicolumn{2}{c}{\textbf{Llama-8B}} & \multicolumn{2}{c}{\textbf{Gemma-4-26B}} & \multicolumn{2}{c}{\textbf{Qwen3-32B}} \\
        \cmidrule(lr){2-3} \cmidrule(lr){4-5} \cmidrule(lr){6-7} \cmidrule(lr){8-9}
        & \textbf{BERTScore} & \textbf{LLM-Judge} & \textbf{BERTScore} & \textbf{LLM-Judge} & \textbf{BERTScore} & \textbf{LLM-Judge} & \textbf{BERTScore} & \textbf{LLM-Judge} \\
        \midrule
        Greek Language    & 68.17 & 77.54 & 69.56 & 56.15 & 72.26 & 75.90 & 70.57 & 74.75 \\
        Mathematics       & 68.15 & 66.89 & 65.65 & 41.06 & 67.56 & 87.02 & 71.22 & 88.91 \\
        Physics           & 71.83 & 69.44 & 75.11 & 47.22 & 79.82 & 75.00 & 77.55 & 69.44 \\
        \midrule
        \textbf{Aggregate} & \textbf{68.31} & \textbf{69.93} & \textbf{67.11} & \textbf{45.48} & \textbf{69.36} & \textbf{83.46} & \textbf{71.29} & \textbf{84.21} \\
        \bottomrule
    \end{tabular}}
    \vspace{-0.2cm}
    \caption{Comparison of evaluation metrics (\%) for zero-shot open-ended tasks in the Prot-Ex benchmark. The LLM-as-a-Judge approach utilizes Gemma-3-27B via the Inspect AI framework.}
    \label{tab:rq2_protipa_metrics}
\end{table*}

\begin{table*}[t]
    \centering
    \resizebox{\textwidth}{!}{%
    \begin{tabular}{l cc cc cc cc}
        \toprule
        \multirow{2}{*}{\textbf{Subject}} & \multicolumn{2}{c}{\textbf{KriKri-8B}} & \multicolumn{2}{c}{\textbf{Llama-8B}} & \multicolumn{2}{c}{\textbf{Gemma-4-26B}} & \multicolumn{2}{c}{\textbf{Qwen3-32B}} \\
        \cmidrule(lr){2-3} \cmidrule(lr){4-5} \cmidrule(lr){6-7} \cmidrule(lr){8-9}
        & \textbf{BERTScore} & \textbf{LLM-Judge} & \textbf{BERTScore} & \textbf{LLM-Judge} & \textbf{BERTScore} & \textbf{LLM-Judge} & \textbf{BERTScore} & \textbf{LLM-Judge} \\
        \midrule
        Ancient Greek     & 64.37 & 57.06 & 67.71 & 35.31 & 67.69 & 65.53 & 61.89 & 64.62 \\
        Economics         & 65.24 & 56.71 & 63.70 & 24.32 & 66.13 & 76.78 & 63.34 & 74.59 \\
        Physics           & 67.06 & 51.37 & 65.77 & 21.23 & 67.96 & 70.55 & 69.38 & 74.66 \\
        History           & 70.17 & 68.98 & 69.61 & 44.44 & 70.33 & 66.20 & 70.43 & 63.33 \\
        Latin             & 55.20 & 51.68 & 62.11 & 32.72 & 62.63 & 64.03 & 53.29 & 66.14 \\
        Mathematics       & 72.47 & 61.47 & 69.94 & 19.47 & 71.67 & 91.58 & 73.88 & 93.42 \\
        Greek Language    & 70.47 & 82.02 & 68.88 & 56.55 & 70.54 & 82.02 & 69.73 & 79.88 \\
        Computer Science  & 64.60 & 71.35 & 65.20 & 41.35 & 64.83 & 77.88 & 65.84 & 86.54 \\
        Chemistry         & 62.70 & 55.33 & 61.04 & 23.17 & 64.29 & 84.47 & 63.79 & 88.50 \\
        Biology           & 67.58 & 63.38 & 68.18 & 30.30 & 68.20 & 69.75 & 62.09 & 70.86 \\
        \midrule
        \textbf{Aggregate} & \textbf{64.77} & \textbf{59.33} & \textbf{65.68} & \textbf{30.81} & \textbf{66.88} & \textbf{74.01} & \textbf{63.86} & \textbf{75.49} \\
        \bottomrule
    \end{tabular}}
    \vspace{-0.2cm}
    \caption{Comparison of evaluation metrics (\%) for zero-shot open-ended tasks in the Pan-Ex benchmark. The LLM-as-a-Judge approach utilizes Gemma-3-27B via the Inspect AI framework.}
    \label{tab:rq2_panellinies_metrics}
\end{table*}

\subsection{BERTScore vs. LLM-as-a-judge}

Tables~\ref{tab:rq2_protipa_metrics} and~\ref{tab:rq2_panellinies_metrics} present zero-shot performance on OEQs using BERTScore and Inspect AI for both benchmarks. Concerning the LLM-as-a-Judge methodology, we utilized Gemma-3-27B-it as the evaluator, guided by subject-specific rubric prompts that assigned a distinct persona (e.g., a strict Greek national examiner) alongside granular grading rules on a 0.0 to 1.0 scale (detailed in Appendix~\ref{sec:appendix_prompts}). 

While this framework provides a nuanced assessment of the models' reasoning capabilities, qualitative analysis revealed slight evaluator leniency.  The judge model occasionally awarded partial credit (0.25) for mere effort on incorrect answers, as seen in a Pan-Ex Ancient Greek task (see Appendix~\ref{subsec:appendix_leniency}). Exploring stricter negative-constraint prompting or alternative judges remains for future work.

Results highlight a notable contrast between traditional lexical metrics and reasoning-based evaluations. BERTScore remains relatively steady across both benchmarks (65\% and 70\%), as it relies primarily on lexical overlap and contextual embeddings rather than on factual correctness or logical flow. Consequently, it often assigns high scores to incorrect outputs; for instance, it awarded a 0.70 to a completely wrong Prot-Ex Mathematics answer, whereas the LLM judge correctly assigned a score of 0.0 (see Appendix~\ref{subsec:appendix_bertscore}).

The data demonstrate a compression effect in BERTScore outputs. BERTScore aggregates range between 64\% and 71\% across models and benchmarks. The metric over-reports the performance of Llama-8B and under-reports the performance of Gemma-4-26B. In the Pan-Ex benchmark, BERTScore evaluates Llama-8B at 65.68\%, while the LLM-Judge evaluates it at 30.81\%. In the Prot-Ex benchmark, BERTScore evaluates Gemma-4-26B at 69.36\%, while the LLM-Judge evaluates it at 83.46\%.


Overall, we observe that Qwen3-32B consistently demonstrated the highest proficiency and accuracy across both benchmarks, achieving aggregate LLM-Judge scores of 84.21\% in Prot-Ex and 75.49\% in Pan-Ex. Notably, the Greek-focused KriKri-8B significantly outperformed its base foundation model, Llama-8B. For instance, it nearly doubled Llama-8B's aggregate score in the Pan-Ex benchmark (59.33\% vs. 30.81\%) and reached 82.02\% in the Pan-Ex Greek Language subject. A representative example occurred in a Prot-Ex Modern Greek task, where KriKri-8B correctly generated the gold answer (scoring 1.0), while Llama-8B yielded a partially accurate response scoring only 0.50 (see Appendix~\ref{subsec:appendix_krikri_llama}). This performance gap extends to STEM disciplines; in the Pan-Ex Mathematics open-ended evaluation, Llama-8B scores 19.47\%, whereas KriKri-8B achieves 61.47\%. This 42-point increase may be attributed to the impact of Greek-specific continued pretraining.

\begin{table*}[t]
    \centering
    \resizebox{\textwidth}{!}{%
    \begin{tabular}{l cc cc cc}
        \toprule
        \multirow{2}{*}{\textbf{Model}} & \multicolumn{2}{c}{\textbf{Closed}} & \multicolumn{2}{c}{\textbf{Structured}} & \multicolumn{2}{c}{\textbf{Open-ended}} \\
        \cmidrule(lr){2-3} \cmidrule(lr){4-5} \cmidrule(lr){6-7}
        & \textbf{Zero-Shot} & \textbf{Few-Shot} & \textbf{Zero-Shot} & \textbf{Few-Shot} & \textbf{Zero-Shot} & \textbf{Few-Shot} \\
        \midrule
        KriKri-8B & 47.10 & 52.20 & 21.67 & 8.61 & 69.93 & 68.03 \\
        Llama-8B  & 43.89 & 49.06 & 7.08 & 8.61 & 45.48 & 44.30 \\
        Gemma-4-26B & 67.90 & 69.43 & 35.28 & 36.94 & 83.46 & 83.64 \\
        Qwen3-32B  & 62.25 & 62.81 & 30.35 & 32.85 & 84.21 & 84.05 \\
        \bottomrule
    \end{tabular}}
    \vspace{-0.2cm}
    \caption{Aggregate impact of few-shot prompting on model performance (\%) across task types in the Prot-Ex benchmark. \textit{Note: The aggregate scores for Structured tasks reflect only the Greek Language subject, as other subjects do not contain questions in this format.}}
    \label{tab:rq3_protipa_fewshot}
\end{table*}

\begin{table*}[t]
    \centering
    \resizebox{\textwidth}{!}{%
    \begin{tabular}{l cc cc cc}
        \toprule
        \multirow{2}{*}{\textbf{Model}} & \multicolumn{2}{c}{\textbf{Closed}} & \multicolumn{2}{c}{\textbf{Structured}} & \multicolumn{2}{c}{\textbf{Open-ended}} \\
        \cmidrule(lr){2-3} \cmidrule(lr){4-5} \cmidrule(lr){6-7}
        & \textbf{Zero-Shot} & \textbf{Few-Shot} & \textbf{Zero-Shot} & \textbf{Few-Shot} & \textbf{Zero-Shot} & \textbf{Few-Shot} \\
        \midrule
        KriKri-8B & 56.74 & 65.28 & 15.57 & 4.88 & 59.33 & 60.13 \\
        Llama-8B  & 49.48 & 56.22 & 11.89 & 0.41 & 30.81 & 32.49 \\
        Gemma-4-26B & 72.54 & 78.76 & 28.70 & 30.77 & 74.01 & 70.34 \\
        Qwen3-32B  & 66.06 & 71.76 & 23.86 & 22.11 & 75.49 & 75.71 \\
        \bottomrule
    \end{tabular}}
    \vspace{-0.2cm}
    \caption{Aggregate impact of few-shot prompting on model performance (\%) across task types in the Pan-Ex benchmark.}
    \label{tab:rq3_panellinies_fewshot}
\end{table*}

\subsection{Performance Across Baseline and Few-shot examples} 

As shown in tables~\ref{tab:rq3_protipa_fewshot} and~\ref{tab:rq3_panellinies_fewshot}, we analyzed the aggregate impact of 5-shot prompting across different task formats, revealing a distinct pattern. It is observed that few-shot examples consistently improved performance in CEQ tasks across all models and both benchmarks. The most notable remark was made in the Pan-Ex benchmark, where KriKri-8B and Llama-8B improved by 8.5 and 6.7 percentage points, respectively. This suggests that providing synthetic examples effectively aligns the models with the expected objective formats (e.g., multiple-choice; see Appendix~\ref{sec:appendix_fewshot}).

Conversely, OEQ tasks exhibited remarkable stability in both benchmarks. The inclusion of few-shot examples yielded minimal performance changes across the board. This indicates that for open-ended generation, zero-shot instructions—combined with a strong system prompt—are largely sufficient for the models to understand the reasoning and formatting requirements, rendering additional context redundant.

Interestingly, SQ tasks experienced a negative impact from few-shot prompting, particularly for the 8B parameter models. KriKri-8B saw a significant drop in both benchmarks (e.g., from 21.67\% to 8.61\% in Prot-Ex), while Llama-8B collapsed almost entirely in Pan-Ex, diving from 11.89\% to a mere 0.41\% (see Appendix~\ref{sec:appendix_llama_collapse}). Notably, larger models like Gemma and Qwen survived this few-shot SQ collapse, validating the capacity overload hypothesis for 8B models. This degradation implies that filling the context window with complex synthetic examples might overwhelm smaller models, causing them to lose track of the specific output constraints or to become distracted by the lengthy and verbose prompt.

\section{Discussion}

Regarding RQ1, performance across both benchmarks heavily depends on task modality and subject, with SQ tasks consistently yielding the lowest scores. Interestingly, larger models (e.g., Qwen3-32B) sometimes excel in open-ended generation (88.91\% in Prot-Ex Mathematics) while struggling with CEQs, likely favoring reasoning articulation over strict option mapping. Furthermore, while larger models predictably dominate STEM, the localized KriKri-8B outperforms them in humanities (Greek Language subject), reaching 96.67\% in Pan-Ex CEQ tasks. This proves that targeted linguistic pre-training can effectively offset lower parameter counts in specialized domains.

Addressing RQ2, traditional lexical metrics like BERTScore fall short for open-ended reasoning, statically measuring lexical overlap rather than factual correctness and reasoning. Conversely, the LLM-as-a-Judge methodology (via Inspect AI) provides a more nuanced and qualitative assessment, though occasional judge leniency towards incorrect answers may compromise evaluation strictness. Overall, Qwen3-32B achieved the highest open-ended aggregate score (84.21\% in Prot-Ex). Meanwhile, the Greek-adapted KriKri-8B consistently received higher evaluations in OEQs than its base model, Llama-8B (a $\sim$20\% Pan-Ex gap), from the Gemma-3-27B judge.

Concerning RQ3, 5-shot prompting primarily benefits CEQs, especially for smaller models (KriKri-8B and Llama-8B), which improved by $\sim$7 percentage points in Pan-Ex. Conversely, OEQs showed minimal fluctuations, indicating that zero-shot instructions and strong system prompts are sufficient for accurate text generation. Notably, few-shot prompting negatively impacted SQs, causing a sharp decline in 8B models; Llama-8B plummeted from 11.89\% to 0.41\% in Pan-Ex, exhibiting erratic behavior in matching and fill-in-the-gaps tasks. This suggests overloading the context window with complex synthetic examples overwhelms smaller models, distracting them from strict output constraints.

\section{Resources}
We release Prot-Ex and Pan-Ex benchmarks as open-source resources on Hugging Face (available at \url{https://huggingface.co/datasets/ilsp/greek-protipa-exams} and \url{https://huggingface.co/datasets/ilsp/panellinies-exams-dataset}, respectively). Additionally, the accompanying codebase, including evaluation scripts and management commands to use the given datasets, is publicly available in our GitHub repositories (\url{https://github.com/PK9811-hub/protipa_exams_dataset} and \url{https://github.com/PK9811-hub/panellinies_exams_dataset}).

\section{Conclusions}
This study demonstrates that while large-scale models predictably excel in complex STEM reasoning, parameter size is not the sole determinant of success. Particularly for a language with limited benchmark coverage such as Greek, smaller but localized models, such as KriKri-8B, can outperform their massive counterparts in linguistically demanding humanities subjects. This highlights that targeted linguistic adaptation and focused domain training can effectively offset lower parameter counts, offering an efficient paradigm for specialized educational applications.

Furthermore, our evaluation exposes the limitations of traditional lexical metrics like BERTScore in capturing factual correctness and logical flow during open-ended reasoning. Adopting an LLM-as-a-Judge methodology provides a more nuanced qualitative assessment. However, this approach is not entirely infallible; we observed instances of evaluator leniency where the judge model awarded partial credit for flawed answers. This underscores that while automated LLM evaluation is superior to static metrics, it requires further refinement via strict negative-constraint prompting.

Finally, our findings reveal that few-shot prompting is not a panacea. While providing synthetic examples clearly benefits CEQs, it yields negligible improvements in OEQs and actively degrades performance in SQ tasks, especially for smaller models. Overloading the context window causes 8B models to lose structural focus, leading to erratic behavior. Ultimately, prompt engineering must be carefully tailored to both the specific task modality and the architectural constraints of each model.

In terms of future work, a key direction involves extending our evaluation paradigm to native Vision Large Language Models (VLMs). Since our current methodology relies on textualized visual contexts, testing VLMs directly on the raw image inputs will allow us to assess their inherent multimodal reasoning capabilities. This will provide critical insights into whether processing visual data 
introduces performance decline or improvements in multimodal QAs compared to text-only alternatives.



\section{Limitations}

Due to the unavailability of source exam data, our study faces certain limitations regarding data composition. The Prot-Ex benchmark contains temporal discontinuities (e.g., missing files or official solutions for the years 2014, 2015, and 2018) and excludes essay-based components, thus preventing the assessment of the models' extensive writing capabilities. Additionally, the Physics subject comprises only 9 questions; consequently, performance metrics for this specific domain lack statistical robustness and should be interpreted with caution.

A persistent challenge in evaluating on educational benchmarks is the potential risk of data contamination. 
Although these original materials were released in noisy, unstructured formats (e.g., raw PDFs and Word documents), making direct memorization of structured question-answer pairs highly unlikely, contamination cannot be entirely ruled out. To address this, we constructed a private test set holdout, withholding the 2019 exams for Prot-Ex and the 2026 exams for Pan-Ex from public releases. This serves as a robust safeguard against evaluation leakage and ensures the integrity of our baseline measurements.

Furthermore, while we observe a severe performance degradation in Structured Questions (SQ) under few-shot settings for smaller models, a detailed ablation study to definitively isolate context-window overload from prompt-format confusion was deferred due to computational budget constraints. We plan to incorporate these extended ablations in the final version of this work.

\section{Ethical Considerations}

The Prot-Ex and Pan-Ex benchmarks comprise content derived from official educational bodies, including the Greek Ministry of Education, the Governing Body of Model and Experimental schools, and the OEFE organization. We explicitly acknowledge that all original exam materials remain the intellectual property of these respective entities. Our use of this data is strictly limited to non-commercial, academic research, in accordance with European text and data mining exceptions for scientific purposes and standard fair use principles. 

From a broader ethical perspective, releasing these resources addresses the ongoing disparity in LLM evaluation, which predominantly focuses on high-resource languages like English. By providing robust benchmarks for Greek, we aim to support the development of linguistically inclusive and unbiased AI models.

While every effort has been made to ensure the accuracy and completeness of these datasets, any errors, omissions, or formatting issues are the result of processing and transformation pipelines and are not related to the original sources.

\bibliography{references}

\newpage
\appendix

\section{Examples with LLM-generated image descriptions and transcriptions}
\label{sec:appendix_multimodal}

\begin{figure}[htbp]
  \begin{subfigure}[b]{\linewidth}
    \fbox{%
      \parbox{\dimexpr\linewidth-2\fboxsep-2\fboxrule\relax}{%
        {\centering \includegraphics[width=0.55\linewidth]{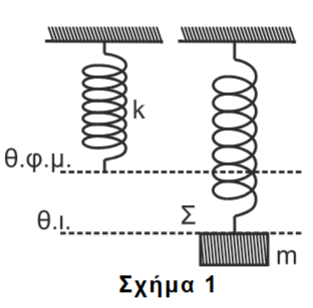}\par}
        \vspace{1mm}

        \small
        \textbf{Description:} Two identical vertical springs with constant $k$ attached to a ceiling. The left spring is at its natural length. The right spring has a block $\Sigma$ of mass $m$ attached to its bottom, extending it. A dotted horizontal line marks the natural length ($\theta.\phi.\mu.$) passing through the bottom of the left spring and a lower dotted horizontal line marks the equilibrium position ($\theta.\iota.$) passing through the top of the mass block attached to the right spring. \\
        \textbf{Transcription:} Σχήμα 1
      }%
    }
    \caption{Pan-Ex (Physics 2022) example.}
    \label{fig:multimodal_physics}
  \end{subfigure}

  \vspace{3mm}

  \begin{subfigure}[b]{\linewidth}
    \fbox{%
      \parbox{\dimexpr\linewidth-2\fboxsep-2\fboxrule\relax}{%
        {\centering \includegraphics[width=0.5\linewidth]{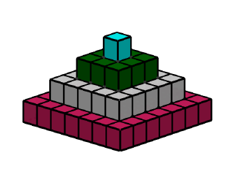}\par}
        \vspace{1mm}
        
        \small
        \textbf{Description:} The image displays a 3D stepped pyramid structure constructed from identical cubes. The structure has four distinct horizontal layers, each a different color and increasing in size from top to bottom.
        \begin{itemize}[nolistsep, leftmargin=*]
          \item Level 1 (Top): A single Cyan cube (Dimensions: $1 \times 1$)
          \item Level 2: Green cubes forming a square platform (Dimensions: $3 \times 3$)
          \item Level 3: Grey cubes forming a square platform (Dimensions: $5 \times 5$)
          \item Level 4 (Bottom): Red/Maroon cubes forming a square platform (Dimensions: $7 \times 7$)
        \end{itemize}
      }%
    }
    \caption{Prot-Ex (Mathematics 2025) example.}
    \label{fig:multimodal_maths}
  \end{subfigure}

  \vspace{2mm}
  \caption{Examples with LLM-generated image descriptions and transcriptions from the Pan-Ex (Physics 2022) and Prot-Ex (Mathematics 2025) benchmarks.}
  \label{fig:multimodal_examples}
\end{figure}

Figure~\ref{fig:multimodal_examples} illustrates two representative examples of the LLM-generated visual context used in our benchmarks. For each image entry, Gemini 3.1 Pro was prompted to produce (i) a detailed textual description of the visual content and (ii) a transcription of any text visible in the image. These textualized representations serve as the sole visual input to the text-only LLMs evaluated in this study, enabling them to reason over image-based questions without native multimodal capabilities.


\section{Prot-Ex and Pan-Ex Public Benchmarks: Subject and Task Format Distribution}
\label{sec:appendix_public_distribution}

\begin{table}[H]
    \centering
    \small
    \begin{tabular}{l r r r r}
        \toprule
        \textbf{Subject} & \textbf{CEQ} & \textbf{SQ} & \textbf{OEQ} & \textbf{Total Items} \\
        \midrule
        \multicolumn{5}{c}{\textbf{Prot-Ex}} \\
        \midrule
        Greek Language    & 735 & 57 & 61  & 853 \\
        Mathematics       & 599 & 0  & 151 & 750 \\
        Religious Studies & 90  & 0  & 0   & 90  \\
        Physics           & 0   & 0  & 9   & 9   \\
        \cmidrule(lr){1-5}
        \textbf{Total} & \textbf{1,424} & \textbf{57} & \textbf{221} & \textbf{1,702} \\
        \midrule
        \multicolumn{5}{c}{\textbf{Pan-Ex}} \\
        \midrule
        Ancient Greek     & 45  & 16 & 131 & 192 \\
        Latin             & 15  & 13 & 149 & 177 \\
        Chemistry         & 52  & 0  & 123 & 175 \\
        Physics           & 74  & 0  & 73  & 147 \\
        Biology           & 37  & 4  & 99  & 140 \\
        Mathematics       & 31  & 0  & 95  & 126 \\
        Economics         & 42  & 0  & 73  & 115 \\
        Computer Science  & 30  & 8  & 52  & 90  \\
        History           & 30  & 0  & 54  & 84  \\
        Greek Language    & 30  & 0  & 42  & 72  \\
        \cmidrule(lr){1-5}
        \textbf{Total} & \textbf{386} & \textbf{41} & \textbf{891} & \textbf{1,318} \\
        \bottomrule
    \end{tabular}
    \caption{Subject-wise distribution across Closed-Ended Questions (CEQ), Structured Questions (SQ), and Open-Ended Questions (OEQ) for the public Prot-Ex and Pan-Ex benchmarks.}
    \label{tab:public_subject_distribution}
\end{table}

\section{Prot-Ex and Pan-Ex Private Test Sets: Subject and Task Format Distribution}

\label{sec:appendix_private_distribution}
\begin{table}[H]
    \centering
    \small
    \begin{tabular}{l r r r r}
        \toprule
        \textbf{Subject} & \textbf{CEQ} & \textbf{SQ} & \textbf{OEQ} & \textbf{Total Items} \\
        \midrule
        \multicolumn{5}{c}{\textbf{Prot-Ex Private}} \\
        \midrule
        Greek Language    & 35 & 5 & 0 & 40 \\
        Mathematics       & 16 & 0 & 8 & 24 \\
        \cmidrule(lr){1-5}
        \textbf{Total} & \textbf{51} & \textbf{5} & \textbf{8} & \textbf{64} \\
        \midrule
        \multicolumn{5}{c}{\textbf{Pan-Ex Private}} \\
        \midrule
        Ancient Greek     & 7  & 0 & 23 & 30 \\
        Latin             & 5  & 7 & 29 & 41 \\
        Chemistry         & 12 & 0 & 20 & 32 \\
        Physics           & 12 & 0 & 11 & 23 \\
        Biology           & 5  & 1 & 11 & 17 \\
        Mathematics       & 5  & 0 & 16 & 21 \\
        Economics         & 7  & 0 & 11 & 18 \\
        Computer Science  & 5  & 2 & 6  & 13 \\
        History           & 5  & 0 & 8  & 13 \\
        Greek Language    & 5  & 0 & 9  & 14 \\
        \cmidrule(lr){1-5}
        \textbf{Total} & \textbf{68} & \textbf{10} & \textbf{144} & \textbf{222} \\
        \bottomrule
    \end{tabular}
    \caption{Subject-wise distribution across Closed-Ended Questions (CEQ), Structured Questions (SQ), and Open-Ended Questions (OEQ) for the private Prot-Ex and Pan-Ex test sets.}
    \label{tab:private_subject_distribution}
\end{table}

\section{Results on the Prot-Ex and Pan-Ex Private Test Sets} 
\label{sec:appendix_private}

This appendix reports detailed model performance metrics on the withheld private test sets for both benchmarks (2019 split for Prot-Ex, $N = 64$; 2026 split for Pan-Ex, $N = 222$). Table~\ref{tab:protex_private_appendix} presents the zero-shot and few-shot results across closed, structured, and open-ended question formats for Prot-Ex, while Table~\ref{tab:panex_private_appendix} details the corresponding performance evaluation on Pan-Ex. Across both private test sets, overall model rankings and relative format behaviors closely mirror the primary public dataset results reported in Section~\ref{sec:results}.

\begin{table}[h!]
    \centering
    \resizebox{\linewidth}{!}{%
    \begin{tabular}{ll ccc}
        \toprule
        \textbf{Model} & \textbf{Setup} & \textbf{Closed} & \textbf{Struct.} & \textbf{Open} \\
        \midrule
        \multirow{2}{*}{KriKri-8B}   & Zero-Shot & 64.71 & 36.00 & 56.25 \\
                                     & Few-Shot  & 60.78 & 56.00 & 84.38 \\
        \cmidrule(lr){1-5}
        \multirow{2}{*}{Llama-8B}    & Zero-Shot & 54.90 & 32.00 & 31.25 \\
                                     & Few-Shot  & 56.86 & 56.00 & 43.75 \\
        \cmidrule(lr){1-5}
        \multirow{2}{*}{Gemma-4-26B} & Zero-Shot & 68.63 & 60.00 & 90.62 \\
                                     & Few-Shot  & 64.71 & 80.00 & 96.88 \\
        \cmidrule(lr){1-5}
        \multirow{2}{*}{Qwen3-32B}   & Zero-Shot & 72.55 & 12.00 & 93.75 \\
                                     & Few-Shot  & 70.59 & 60.00 & 96.88 \\
        \bottomrule
    \end{tabular}}
    \caption{Evaluation results for the private Prot-Ex (2019) test set. Scores represent aggregate accuracy for Closed and Structured question formats under the LM-Eval framework, alongside Inspect AI evaluations for Open-Ended questions.}
    \label{tab:protex_private_appendix}
\end{table}

\begin{table}[h!]
    \centering
    \resizebox{\linewidth}{!}{%
    \begin{tabular}{ll ccc}
        \toprule
        \textbf{Model} & \textbf{Setup} & \textbf{Closed} & \textbf{Struct.} & \textbf{Open} \\
        \midrule
        \multirow{2}{*}{KriKri-8B}   & Zero-Shot & 72.06 & 13.00 & 57.26 \\
                                     & Few-Shot  & 79.41 & 0.00  & 56.18 \\
        \cmidrule(lr){1-5}
        \multirow{2}{*}{Llama-8B}    & Zero-Shot & 57.35 & 6.50  & 31.08 \\
                                     & Few-Shot  & 63.24 & 0.00  & 29.34 \\
        \cmidrule(lr){1-5}
        \multirow{2}{*}{Gemma-4-26B} & Zero-Shot & 80.88 & 23.50 & 73.02 \\
                                     & Few-Shot  & 79.41 & 26.00 & 68.68 \\
        \cmidrule(lr){1-5}
        \multirow{2}{*}{Qwen3-32B}   & Zero-Shot & 79.41 & 24.00 & 75.14 \\
                                     & Few-Shot  & 77.94 & 16.00 & 78.23 \\
        \bottomrule
    \end{tabular}}
    \caption{Evaluation results for the private Pan-Ex (2026) test set. Scores represent aggregate accuracy for Closed and Structured question formats under the LM-Eval framework, alongside Inspect AI evaluations for Open-Ended questions.}
    \label{tab:panex_private_appendix}
\end{table}

\section{LLM-as-a-Judge Evaluation Prompts}
\label{sec:appendix_prompts}

In this section, we provide representative examples of the system instructions and grading rubrics utilized for the LLM-as-a-Judge evaluation methodology. For each prompt, we present the original Greek text provided to the model, followed by its English translation.

\subsection{Modern Greek Language (Pan-Ex)}
\label{subsec:prompts_greek}

\textbf{System Instruction (Original Greek):} \\
\textit{Είσαι ένας 18χρονος τελειόφοιτος Λυκείου που απαντά σε διαγώνισμα Πανελλαδικών στη Νεοελληνική Γλώσσα και Λογοτεχνία. Απάντησε στο ερώτημα συγκροτημένα, με πλούσιο λεξιλόγιο, σωστή δομή, άρτια γραμματική και συντακτικό.}

\vspace{0.2cm}
\noindent\textbf{System Instruction (English Translation):} \\
\textit{You are an 18-year-old high school senior taking the Panhellenic national exam in Modern Greek Language and Literature. Answer the question coherently, with rich vocabulary, proper structure, and flawless grammar and syntax.}

\vspace{0.3cm}
\noindent\textbf{Rubric (Original Greek):} \\
\textit{Είσαι ένας αυστηρός Έλληνας βαθμολογητής Πανελλαδικών Εξετάσεων που διορθώνει το γραπτό Νεοελληνικής Γλώσσας ενός 18χρονου τελειόφοιτου Λυκείου. Αξιολόγησε την απάντηση του μαθητή (Submission) συγκρίνοντάς τη με την πρότυπη λύση (Criterion). Χρησιμοποίησε κλίμακα βαθμολόγησης: 0.0, 0.25, 0.5, 0.75, ή 1.0. Κανόνες:}
\begin{itemize}
    \item \textit{1. Εστίασε στην ορθογραφία, τη γραμματική, το συντακτικό, την ακρίβεια του λεξιλογίου και την πλήρη απόδοση του νοήματος.}
    \item \textit{2. Δώσε 1.0 αν η απάντηση είναι άψογη νοηματικά και συντακτικά, πλήρως τεκμηριωμένη και στοχευμένη.}
    \item \textit{3. Δώσε 0.75 αν βρήκε το σωστό νόημα, αλλά έκανε κάποιο ελαφρύ εκφραστικό, συντακτικό ή ορθογραφικό λάθος.}
    \item \textit{4. Δώσε 0.50 αν βρήκε μέρος της απάντησης ή αν η διατύπωση είναι ασαφής, άκομψη ή δημιουργεί πλεονασμούς.}
    \item \textit{5. Δώσε 0.25 αν η απάντηση είναι ελλιπής ή μερικώς εκτός θέματος, αλλά περιέχει τουλάχιστον ένα σωστό σημείο αναφοράς.}
    \item \textit{6. Δώσε 0.0 αν η απάντηση είναι εντελώς εκτός θέματος, λανθασμένη ή παρουσιάζει σοβαρότατα πραγματολογικά λάθη.}
\end{itemize}

\noindent\textbf{Rubric (English Translation):} \\
\textit{You are a strict Greek national examiner grading the Modern Greek Language exam paper of an 18-year-old high school senior. Evaluate the student's answer (Submission) by comparing it with the gold standard solution (Criterion). Use the following grading scale: 0.0, 0.25, 0.5, 0.75, or 1.0. Rules:}
\begin{itemize}
    \item \textit{1. Focus on spelling, grammar, syntax, vocabulary accuracy, and the complete rendering of the meaning.}
    \item \textit{2. Provide a 1.0 score if the answer is conceptually and syntactically flawless, fully substantiated, and targeted.}
    \item \textit{3. Provide a 0.75 score if the correct meaning is captured, but there is a minor expressive, syntactic, or spelling error.}
    \item \textit{4. Provide a 0.50 score if part of the answer is correct or if the phrasing is vague, awkward, or creates redundancies.}
    \item \textit{5. Provide a 0.25 score if the answer is incomplete or partially off-topic, but contains at least one correct reference point.}
    \item \textit{6. Provide a 0.0 score if the answer is completely off-topic, incorrect, or presents severe factual errors.}
\end{itemize}

\vspace{0.1cm}

\subsection{Mathematics (Prot-Ex)}
\label{subsec:prompts_math}

\textbf{System Instruction (Original Greek):} \\
\textit{Είσαι ένας 12χρονος Έλληνας μαθητής που απαντά σε διαγώνισμα Μαθηματικών. Λύσε το πρόβλημα βήμα-βήμα, δείχνοντας τις πράξεις σου απλά, και γράψε το τελικό αριθμητικό αποτέλεσμα καθαρά στο τέλος.}

\vspace{0.2cm}
\noindent\textbf{System Instruction (English Translation):} \\
\textit{You are a 12-year-old Greek student taking a Mathematics exam. Solve the problem step-by-step, showing your operations simply, and write the final numerical result clearly at the end.}

\vspace{0.3cm}
\noindent\textbf{Rubric (Original Greek):} \\
\textit{Είσαι ένας αυστηρός Έλληνας εκπαιδευτικός που βαθμολογεί το γραπτό Μαθηματικών ενός 12χρονου μαθητή. Αξιολόγησε την απάντηση του μαθητή (Submission) συγκρίνοντάς τη με την πρότυπη λύση (Criterion). Χρησιμοποίησε κλίμακα βαθμολόγησης: 0.0, 0.25, 0.5, 0.75, ή 1.0. Κανόνες:}
\begin{itemize}
    \item \textit{1. Δώσε 1.0 αν η μεθοδολογία είναι σωστή και το τελικό αποτέλεσμα ταυτίζεται απόλυτα με το Criterion.}
    \item \textit{2. Δώσε 0.75 αν η μεθοδολογία είναι ολόσωστη αλλά υπάρχει ένα μικρό αριθμητικό λάθος στο τελικό αποτέλεσμα.}
    \item \textit{3. Δώσε 0.50 αν ο μαθητής ακολούθησε τα σωστά βήματα μέχρι τη μέση ή βρήκε μόνο μέρος της λύσης (π.χ. τη μία από τις δύο λύσεις μιας εξίσωσης).}
    \item \textit{4. Δώσε 0.25 αν η μεθοδολογία είναι λανθασμένη ή ατελής, αλλά εφάρμοσε σωστά κάποιον βασικό τύπο ή έκανε μια σωστή αρχική σκέψη.}
    \item \textit{5. Δώσε 0.0 αν και η λογική και το αποτέλεσμα είναι εντελώς λανθασμένα ή δεν υπάρχει καμία προσπάθεια λύσης.}
\end{itemize}

\noindent\textbf{Rubric (English Translation):} \\
\textit{You are a strict Greek educator grading the Mathematics exam paper of a 12-year-old student. Evaluate the student's answer (Submission) by comparing it with the gold standard solution (Criterion). Use the following grading scale: 0.0, 0.25, 0.5, 0.75, or 1.0. Rules:}
\begin{itemize}
    \item \textit{1. Provide a 1.0 score if the methodology is correct and the final result matches the Criterion perfectly.}
    \item \textit{2. Provide a 0.75 score if the methodology is entirely correct but there is a minor arithmetic error in the final result.}
    \item \textit{3. Provide a 0.50 score if the student followed the correct steps halfway or found only part of the solution (e.g., one of two solutions to an equation).}
    \item \textit{4. Provide a 0.25 score if the methodology is incorrect or incomplete, but the student correctly applied a basic formula or made a correct initial reasoning step.}
    \item \textit{5. Provide a 0.0 score if both the logic and the result are completely incorrect, or if there is no attempt to solve the problem.}
\end{itemize}

\section{LLM-as-a-Judge Examples from Evaluation Logs}
\label{sec:appendix_logs}

\subsection{Evaluator Leniency: Pan-Ex Ancient Greek}
\label{subsec:appendix_leniency}

This example demonstrates a case where the judge model (Gemma-3-27B-it) awarded partial credit to a model (Gemma-4-26B) for an incorrect answer. The judge recognized the student's effort to explain the broader context, despite the failure to address the specific grammatical question.

\vspace{0.2cm}
\noindent \textbf{Question ID:} \texttt{ancient\_greek\_gel\_2020\_Α1.β.1}

\vspace{0.2cm}
\noindent \textbf{Question:} \\
\textit{«σφεῖς ἐντιθέναι...»: To which word in the ancient text does the pronoun «σφεῖς» refer?}

\vspace{0.2cm}
\noindent \textbf{Target:} \\
\textit{The word in the ancient text to which the pronoun «σφεῖς» refers is: «ἐπαγγελλόμενοί».}

\vspace{0.2cm}
\noindent \textbf{Student Answer (Gemma-4-26B):} \\
\textit{The pronoun «σφεῖς» (meaning "only", "simply") refers to the concept of \textbf{ἐπιστήμης} [knowledge/science]. In this specific part of the text, Plato reproduces the view of certain people who argue that education is not the actual acquisition of knowledge, but simply the addition of a "science" to the soul, without it residing within it, exactly as adding sight to blind eyes does not make them capable of vision.}

\vspace{0.2cm}
\noindent \textbf{Judge Explanation (Gemma-3-27B-it):} \\
\textit{The student's answer incorrectly interprets the reference of the pronoun "σφεῖς". Instead of identifying the word it refers to (i.e., "ἐπαγγελλόμενοί"), it focuses on the concept of the word "ἐπιστήμης" and attempts an interpretation of the excerpt. The understanding of the excerpt is partially correct, but the answer does not focus on what the question asked. Therefore, it receives a low grade.}

\vspace{0.2cm}
\noindent \textbf{Score:} 0.25

\subsection{Metric Discrepancy: Prot-Ex Mathematics}
\label{subsec:appendix_bertscore}

This example highlights the discrepancy between lexical metrics and reasoning-based evaluation. The student model (Llama-8B) provides a completely incorrect final answer without demonstrating any logical steps. While the judge model correctly identifies the error and assigns a score of 0.0, BERTScore assigns a high score of $\sim$0.71 due to high vocabulary and numerical overlap with the target solution.

\vspace{0.2cm}
\noindent \textbf{Question ID:} \texttt{math\_gym\_2016\_1\_10.2}

\vspace{0.2cm}
\noindent \textbf{Question:} \\
\textit{Orpheus and Hypatia have a total of 27 candies together. Orpheus gave some of his candies to Hypatia. Hypatia now has twice as many as she had, while Orpheus now has three more than Hypatia. b) How many candies did each have initially?}

\vspace{0.2cm}
\noindent \textbf{Target:} \\
\textit{12 : 2 = 6 candies for Hypatia, 27 - 6 = 21 candies for Orpheus. \\
Alternatively (for a and b): \\
2 parts Hypatia and 2 parts + 3 candies Orpheus. \\
27 - 3 = 24, 24 : 4 = 6. \\
Now: Hypatia $2 \cdot 6 = 12$ candies and Orpheus $2 \cdot 6 + 3 = 15$ candies. \\
Initially: Hypatia 6 candies and Orpheus 27 - 6 = 21 candies.}

\vspace{0.2cm}
\noindent \textbf{Student Answer (Llama-8B):} \\
\textit{Answer: 15 candies for Orpheus and 12 candies for Hypatia.}

\vspace{0.2cm}
\noindent \textbf{Judge Explanation (Gemma-3-27B-it):} \\
\textit{The answer is completely incorrect. The student does not seem to have understood the description of the problem and has not followed any logical path to find the solution. The answer 15 and 12 does not satisfy any of the conditions given in the problem.}

\vspace{0.2cm}
\noindent \textbf{LLM Judge Score:} 0.0 \\
\textbf{BERTScore:} 0.71

\subsection{Model Comparison: Prot-Ex Modern Greek}
\label{subsec:appendix_krikri_llama}

This example illustrates the performance gap between the Greek-adapted KriKri-8B and its base foundation model, Llama-8B. While KriKri-8B correctly identifies a correct synonym that fits the semantic context, Llama-8B struggles with the nuances of the Greek language, selecting a contextually inappropriate verb that alters the meaning of the sentence, resulting in a lower score from the LLM-as-a-Judge (Gemma-3-27B-it).

\vspace{0.2cm}
\noindent \textbf{Question ID:} \texttt{greek\_language\_gym\_2016\_1\_2.4}

\vspace{0.2cm}
\noindent \textbf{Question:} \\
\textit{In the following phrase, replace the underlined word with another word without changing the meaning: "The operating hours of 'Hellenic Cosmos' may \textbf{change} [αλλάξει] without notice."}

\vspace{0.2cm}
\noindent \textbf{Target:} \\
\textit{The operating hours of 'Hellenic Cosmos' may be altered/modified [μεταβληθεί/τροποποιηθεί] without notice.}

\vspace{0.3cm}
\noindent \textbf{Student Answer (KriKri-8B):} \\
\textit{The phrase with the replacement of the underlined word is: "The operating hours of 'Hellenic Cosmos' may be \textbf{modified} [τροποποιηθεί] without notice." The word "modified" maintains the meaning of the possibility of altering/modifying the schedule without prior notice.}

\vspace{0.2cm}
\noindent \textbf{Judge Explanation for KriKri-8B:} \\
\textit{The answer is flawless. The student selected the correct synonym ("τροποποιηθεί") for the underlined word, fully maintaining the meaning of the sentence. There are no spelling or syntax errors. Excellent job!}

\vspace{0.2cm}
\noindent \textbf{Score (KriKri-8B):} 1.0

\vspace{0.3cm}
\noindent \textbf{Student Answer (Llama-8B):} \\
\textit{Answer: "The operating hours of 'Hellenic Cosmos' may change without notice." \\
Replacement: "The operating hours of 'Hellenic Cosmos' may \textbf{transit/move} [μεταβεί] without notice." \\
Explanation: The word "transit/move" means to change location, time, etc., and is appropriate for the required case.}

\vspace{0.2cm}
\noindent \textbf{Judge Explanation for Llama-8B:} \\
\textit{The answer, although attempting to provide an alternative word, is not the most appropriate. The word "μεταβεί" [transit/move] has a different nuance and does not exactly convey the meaning of changing the schedule. The use of words like "μεταβληθεί" or "τροποποιηθεί" [altered/modified] would be preferable. Therefore, the answer is part of the solution, but not fully satisfactory.}

\vspace{0.2cm}
\noindent \textbf{Score (Llama-8B):} 0.50

\section{Synthetic Examples used in Few-Shot Prompt}
\label{sec:appendix_fewshot}

\subsection{Prot-Ex: Modern Greek Language (Matching)}
\label{subsec:fewshot_protex_greek}

\vspace{0.2cm}
\noindent \textbf{Question (Original Greek):} \\
\textit{Κείμενο 2: Το όνειρο του Άρη. Ο Άρης είναι το βασικό στήριγμα στην τετραμελή οικογένειά του. Οι γονείς του διατηρούν έναν παραδοσιακό φούρνο στο χωριό και καθημερινά αναλαμβάνει τον ρόλο του ταμία για να τους εξυπηρετεί. Φροντίζει ενεργά για τις παραδόσεις των παραγγελιών, μιλάει με ευγένεια στους πελάτες και, παράλληλα, πηγαίνει στις προπονήσεις του. Εκεί, ο προπονητής του αντιλαμβάνεται τις δυνατότητές του στις ταχύτητες, τον παροτρύνει να ενταχθεί στην τοπική ομάδα και σιγά-σιγά τον καθοδηγεί να βελτιώσει τους χρόνους του. Καθώς οι επιδόσεις του εξελίσσονται, ο προπονητής τού παρουσιάζει μία διαφορετική πρόταση που δεν είχε τολμήσει να σκεφτεί ποτέ. Του προτείνεται να συμμετάσχει στο πανελλήνιο πρωτάθλημα στην Αθήνα, όπου η διάκριση θα μπορούσε να του προσφέρει μια θέση σε μεγάλο σύλλογο και μια λαμπρή καριέρα στον αθλητισμό. Ο Άρης θέλει να κάνει το όνειρό του πραγματικότητα, αλλά δε νιώθει έτοιμος να αποχωριστεί τους δικούς του, που βασίζονται τόσο πολύ πάνω του. \\
Γράψε τις φράσεις (1-5) στη στήλη (Α-Γ) στην οποία ταιριάζει η καθεμιά, σύμφωνα με το κείμενο 2: \\
Α. Οικογενειακή επιχείρηση \\
Β. Αθλητική δραστηριότητα \\
Γ. Μελλοντική σταδιοδρομία \\
1. αναλαμβάνει τον ρόλο του ταμία \\
2. Φροντίζει ενεργά για τις παραδόσεις των παραγγελιών \\
3. τον παροτρύνει να ενταχθεί στην τοπική ομάδα \\
4. λαμπρή καριέρα στον αθλητισμό \\
5. δε νιώθει έτοιμος να αποχωριστεί τους δικούς του}

\vspace{0.2cm}
\noindent \textbf{Target:} \\
\textit{A-1, A-2, A-5, B-3, Γ-4}

\vspace{0.3cm}
\noindent \textbf{Question (English Translation):} \\
\textit{Text 2: Aris's dream. Aris is the main pillar of his four-member family. His parents run a traditional bakery in the village, and every day he takes on the role of cashier to help them. He actively takes care of order deliveries, speaks politely to customers, and, at the same time, goes to his training sessions. There, his coach recognizes his potential in sprinting, encourages him to join the local team, and gradually guides him to improve his times. As his performance evolves, the coach presents him with a different proposal he had never dared to think about. He is suggested to participate in the national championship in Athens, where a distinction could offer him a position in a major club and a brilliant career in sports. Aris wants to make his dream come true, but he doesn't feel ready to part with his family, who rely on him so much. \\
Write the phrases (1-5) in the column (A-C) they match, according to Text 2: \\
A. Family business \\
B. Sports activity \\
C. Future career \\
1. takes on the role of cashier \\
2. actively takes care of order deliveries \\
3. encourages him to join the local team \\
4. brilliant career in sports \\
5. doesn't feel ready to part with his family}

\vspace{0.2cm}
\noindent \textbf{Target (English Translation):} \\
\textit{A-1, A-2, A-5, B-3, C-4}

\subsection{Prot-Ex: Mathematics (Multiple Choice)}
\label{subsec:fewshot_protex_math}

\vspace{0.2cm}
\noindent \textbf{Question (Original Greek):} \\
\textit{Η Μαρία στα διαγωνίσματα της Ιστορίας έχει πάρει τις εξής βαθμολογίες: 14, 17, 15, 16. Πόσο πρέπει να πάρει στο 5ο διαγώνισμα για να βγάλει μέσο όρο 16; \\
Α. 15, Β. 16, Γ. 18, Δ. 19, Ε. 20}

\vspace{0.2cm}
\noindent \textbf{Target:} \\
\textit{Γ}

\vspace{0.3cm}
\noindent \textbf{Question (English Translation):} \\
\textit{Maria has received the following grades in her History exams: 14, 17, 15, 16. What score must she get on the 5th exam to achieve an average of 16? \\
A. 15, B. 16, C. 18, D. 19, E. 20}

\vspace{0.2cm}
\noindent \textbf{Target (English Translation):} \\
\textit{C}

\subsection{Pan-Ex: Ancient Greek (Fill in the gaps)}
\label{subsec:fewshot_panex_ancient}

\vspace{0.2cm}
\noindent \textbf{Question (Original Greek):} \\
\textit{Να συμπληρώσετε τις παρακάτω περιόδους λόγου με ουσιαστικά ετυμολογικά συγγενή (απλά ή σύνθετα) της μετοχής «λαμβάνοντας» ώστε να ολοκληρωθεί σωστά το νόημά τους: Η ......... του νέου εργαστηριακού εξοπλισμού θα γίνει την ερχόμενη Δευτέρα.}

\vspace{0.2cm}
\noindent \textbf{Target:} \\
\textit{παραλαβή}

\vspace{0.3cm}
\noindent \textbf{Question (English Translation):} \\
\textit{Fill in the following sentences with nouns etymologically related (simple or compound) to the participle "λαμβάνοντας" (receiving) so that their meaning is correctly completed: The ......... of the new laboratory equipment will take place next Monday.}

\vspace{0.2cm}
\noindent \textbf{Target (English Translation):} \\
\textit{παραλαβή (receipt)}

\subsection{Pan-Ex: Computer Science (Fill in the gaps)}
\label{subsec:fewshot_panex_cs}

\vspace{0.2cm}
\noindent \textbf{Question (Original Greek):} \\
\textit{Δίνεται τετραγωνικός πίνακας ακεραίων A[50, 50]. Το παρακάτω τμήμα αλγορίθμου ελέγχει αν ο πίνακας είναι συμμετρικός ως προς την κύρια διαγώνιό του (δηλαδή αν για κάθε στοιχείο του ισχύει A[i, j] = A[j, i]) χρησιμοποιώντας μια λογική μεταβλητή. Αν βρεθεί έστω και ένα ζευγάρι στοιχείων που να παραβιάζει αυτή τη συνθήκη, η διαδικασία του ελέγχου διακόπτεται. \\
Να γράψετε στο τετράδιό σας τους αριθμούς (1) έως (5) που αντιστοιχούν στα κενά του τμήματος αλγορίθμου και δίπλα ό,τι πρέπει να συμπληρωθεί, έτσι ώστε να επιτελεί τη λειτουργία που περιγράφηκε. \\
\\
Συμμετρικός $\leftarrow$ ...(1)... \\
i $\leftarrow$ 2 \\
ΟΣΟ i $\leq$ 50 ΚΑΙ Συμμετρικός = ...(2)... ΕΠΑΝΑΛΑΒΕ \\
j $\leftarrow$ 1 \\
ΟΣΟ j $<$ ...(3)... ΚΑΙ Συμμετρικός = ΑΛΗΘΗΣ ΕΠΑΝΑΛΑΒΕ \\
ΑΝ A[i, j] $\neq$ A[...(4)...] ΤΟΤΕ \\
Συμμετρικός $\leftarrow$ ...(5)... \\
ΑΛΛΙΩΣ \\
j $\leftarrow$ j + 1 \\
ΤΕΛΟΣ\_ΑΝ \\
ΤΕΛΟΣ\_ΕΠΑΝΑΛΗΨΗΣ \\
i $\leftarrow$ i + 1 \\
ΤΕΛΟΣ\_ΕΠΑΝΑΛΗΨΗΣ}

\vspace{0.2cm}
\noindent \textbf{Target:} \\
\textit{(1) ΑΛΗΘΗΣ, (2) ΑΛΗΘΗΣ, (3) i, (4) j, i, (5) ΨΕΥΔΗΣ}

\vspace{0.3cm}
\noindent \textbf{Question (English Translation):} \\
\textit{An integer square matrix A[50, 50] is given. The following algorithm snippet checks if the matrix is symmetric with respect to its main diagonal (i.e., if for every element A[i, j] = A[j, i]) using a boolean variable. If even one pair of elements violates this condition, the checking process stops. \\
Write in your notebook the numbers (1) to (5) corresponding to the blanks in the algorithm snippet and next to them what needs to be filled in, so that it performs the described function. \\
\\
Symmetric $\leftarrow$ ...(1)... \\
i $\leftarrow$ 2 \\
WHILE i $\leq$ 50 AND Symmetric = ...(2)... DO \\
j $\leftarrow$ 1 \\
WHILE j $<$ ...(3)... AND Symmetric = TRUE DO \\
IF A[i, j] $\neq$ A[...(4)...] THEN \\
Symmetric $\leftarrow$ ...(5)... \\
ELSE \\
j $\leftarrow$ j + 1 \\
END\_IF \\
END\_WHILE \\
i $\leftarrow$ i + 1 \\
END\_WHILE}

\vspace{0.2cm}
\noindent \textbf{Target (English Translation):} \\
\textit{(1) TRUE, (2) TRUE, (3) i, (4) j, i, (5) FALSE}

\section{Few-Shot Degradation in Structured Tasks (Llama-8B)}
\label{sec:appendix_llama_collapse}

\subsection{Pan-Ex Latin}
This example demonstrates the model completely losing its output language consistency. Despite the prompt and all 5 few-shot examples being strictly in Greek, the model hallucinates and answers in English. Furthermore, it incorrectly identifies the grammatical case (identifying it as genitive instead of dative).

\vspace{0.2cm}
\noindent \textbf{Question ID:} \texttt{latin\_gel\_2025\_Δ1.α.3}

\vspace{0.2cm}
\noindent \textbf{Question (Original Greek):} \\
\textit{Να αναγνωρίσετε συντακτικά τον παρακάτω όρο: ancillae: είναι ................. στο .................}

\vspace{0.2cm}
\noindent \textbf{Target:} \\
\textit{έμμεσο αντικείμενο στο ρήμα credidi}

\vspace{0.2cm}
\noindent \textbf{Model Output (Llama-8B 5-shot):} \\
\textit{genitive}

\vspace{0.3cm}
\noindent \textbf{Question (English Translation):} \\
\textit{Syntactically identify the following term: ancillae: it is ................. to the .................}

\vspace{0.2cm}
\noindent \textbf{Target (English Translation):} \\
\textit{indirect object to the verb credidi}

\vspace{0.2cm}
\noindent \textbf{Model Output (English):} \textit{genitive}

\subsection{Pan-Ex Biology}
In this matching task, the model fails to process the logical constraints of the prompt. Instead of correctly mapping the items, it falls into a repetitive generation loop (babbling), assigning the choice "γ" to almost all items in the array.

\vspace{0.2cm}
\noindent \textbf{Question ID:} \texttt{biology\_gel\_2021\_B1}

\vspace{0.2cm}
\noindent \textbf{Target Format:} \\
\textit{["1-α", "2-γ", "3-β", "4-α", "5-γ", "6-β", "7-α"]}

\vspace{0.2cm}
\noindent \textbf{Model Output (Llama-8B 5-shot):} \\
\textit{["1-γ", "2-β", "3-α", "4-γ", "5-γ", "6-γ", "7-γ"]}

\subsection{Pan-Ex Ancient Greek}
In this fill-in-the-gaps task, the model is explicitly instructed to provide a word etymologically related to the participle "φέροντας" (carrying/bearing). Overwhelmed by the few-shot examples in the context, it completely ignores this primary semantic constraint and simply guesses a contextually plausible but unrelated word ("φόβο" instead of "φέρσιμο").

\vspace{0.2cm}
\noindent \textbf{Question ID:} \texttt{ancient\_greek\_gel\_2025\_B3.4}

\vspace{0.2cm}
\noindent \textbf{Question (Original Greek):} \\
\textit{Να συμπληρώσετε τις παρακάτω περιόδους λόγου με ουσιαστικά ετυμολογικά συγγενή (απλά ή σύνθετα) της μετοχής «φέροντας» ώστε να ολοκληρωθεί σωστά το νόημά τους: Με στενοχώρησε το .......... του. Ήταν φοβερά προσβλητικό.}

\vspace{0.2cm}
\noindent \textbf{Target:} \\
\textit{φέρσιμο}

\vspace{0.2cm}
\noindent \textbf{Model Output (Llama-8B 5-shot):} \\
\textit{φόβο}

\vspace{0.3cm}
\noindent \textbf{Question (English Translation):} \\
\textit{Fill in the following sentences with nouns etymologically related (simple or compound) to the participle "φέροντας" (bearing/carrying) so that their meaning is correctly completed: His .......... saddened me. It was terribly offensive.}

\vspace{0.2cm}
\noindent \textbf{Target (English Translation):} \\
\textit{φέρσιμο (behavior)}

\vspace{0.2cm}
\noindent \textbf{Model Output (English Translation):} \\
\textit{φόβο (fear)}

\twocolumn 

\end{document}